\documentclass[sigconf]{acmart}

\copyrightyear{2024}
\acmYear{2024}
\setcopyright{acmlicensed}
\acmConference[MM '24] {Proceedings of the 32nd ACM International Conference on Multimedia}{October 28--November 1, 2024}{Melbourne, VIC, Australia.}
\acmBooktitle{Proceedings of the 32nd ACM International Conference on Multimedia (MM '24), October 28--November 1, 2024, Melbourne, VIC, Australia}
\acmISBN{979-8-4007-0686-8/24/10}
\acmDOI{10.1145/3664647.3680679}
\usepackage{multirow}
\usepackage{balance}
\usepackage{algorithm}
\usepackage{algorithmic}
\usepackage{color}

\newcommand{\ie}{\textit{i.e.}}
\newcommand{\etc}{\textit{etc.}}

\DeclareMathOperator*{\argmin}{arg\,min}
\definecolor{grey}{RGB}{150,150,150} 

\begin{document}

\title{Ego3DT: Tracking Every 3D Object in Ego-centric Videos}

\author{Shengyu Hao}
\authornote{Authors contributed equally to this research.}
\email{shengyuhao@zju.edu.cn}
\affiliation{%
  \department{College of Computer Science and Technology}
  \institution{Zhejiang University}
  \city{Hangzhou}
  \country{China}}

\author{Wenhao Chai}
\authornotemark[1]
\email{wchai@uw.edu}
\affiliation{%
  \institution{University of Washington}
  \city{Seattle}
  \country{USA}
}

\author{Zhonghan Zhao}
\authornotemark[1]
\email{zhaozhonghan@zju.edu.cn}
\affiliation{%
  \department{College of Computer Science and Technology}
  \institution{Zhejiang University}
  \city{Hangzhou}
  \country{China}
}

\author{Meiqi Sun}
\email{meiqi.22@intl.zju.edu.cn}
\affiliation{%
  \department{Zhejiang University-University of Illinois Urbana Champaign Institute}
  \institution{Zhejiang University}
  \city{Haining}
  \country{China}
}

\author{Wendi Hu}
\email{3200105651@zju.edu.cnn}
\affiliation{%
  \department{College of Computer Science and Technology}
  \institution{Zhejiang University}
  \city{Hangzhou}
  \country{China}
}

\author{Jieyang Zhou}
\email{jzhou103@illinois.edu}
\affiliation{%
  \department{Zhejiang University-University of Illinois Urbana Champaign Institute}
  \institution{Zhejiang University}
  \city{Haining}
  \country{China}
}

\author{Yixian Zhao}
\email{3230111486@zju.edu.cn}
\affiliation{%
  \department{Zhejiang University-University of Illinois Urbana Champaign Institute}
  \institution{Zhejiang University}
  \city{Haining}
  \country{China}
}

\author{Qi Li}
\email{3230114803@zju.edu.cn}
\affiliation{%
  \department{Zhejiang University-University of Illinois Urbana Champaign Institute}
  \institution{Zhejiang University}
  \city{Haining}
  \country{China}
}

\author{Yizhou Wang}
\email{ywang26@uw.edu}
\affiliation{%
  \institution{University of Washington}
  \city{Seattle}
  \country{USA}
}

\author{Xi Li}
\email{xilizju@zju.edu.cn}
\authornote{Corresponding author.}
\affiliation{%
  \department{College of Computer Science and Technology}
  \institution{Zhejiang University}
  \city{Hangzhou}
  \country{China}
}

\author{Gaoang Wang}
\authornotemark[2]
\email{gaoangwang@intl.zju.edu.cn}
\affiliation{%
  \department{Zhejiang University-University of Illinois Urbana Champaign Institute}
  \department{College of Computer Science and Technology}
  \institution{Zhejiang University}
  \city{Haining}
  \country{China}
}

\renewcommand{\shortauthors}{Shengyu Hao et al.}

\begin{abstract}
The growing interest in embodied intelligence has brought ego-centric perspectives to contemporary research. One significant challenge within this realm is the accurate localization and tracking of objects in ego-centric videos, primarily due to the substantial variability in viewing angles. Addressing this issue, this paper introduces a novel zero-shot approach for the 3D reconstruction and tracking of all objects from the ego-centric video. We present Ego3DT, a novel framework that initially identifies and extracts detection and segmentation information of objects within the ego environment. Utilizing information from adjacent video frames, Ego3DT dynamically constructs a 3D scene of the ego view using a pre-trained 3D scene reconstruction model. Additionally, we have innovated a dynamic hierarchical association mechanism for creating stable 3D tracking trajectories of objects in ego-centric videos. Moreover, the efficacy of our approach is corroborated by extensive experiments on two newly compiled datasets, with $1.04 \times$ - $2.90\times$ in HOTA, showcasing the robustness and accuracy of our method in diverse ego-centric scenarios.
\end{abstract}


\begin{CCSXML}
<ccs2012>
<concept>
<concept_id>10010147.10010178.10010224.10010225.10010227</concept_id>
<concept_desc>Computing methodologies~Scene understanding</concept_desc>
<concept_significance>300</concept_significance>
</concept>
</ccs2012>
\end{CCSXML}

\ccsdesc[300]{Computing methodologies~Scene understanding}


% \keywords{Do, Not, Us, This, Code, Put, the, Correct, Terms, for,

\keywords{3D Vision, Open Vocabulary Tracking, Ego-centric Videos}



\maketitle

\begin{figure}[ht]
\centering
    \includegraphics[width=\linewidth]{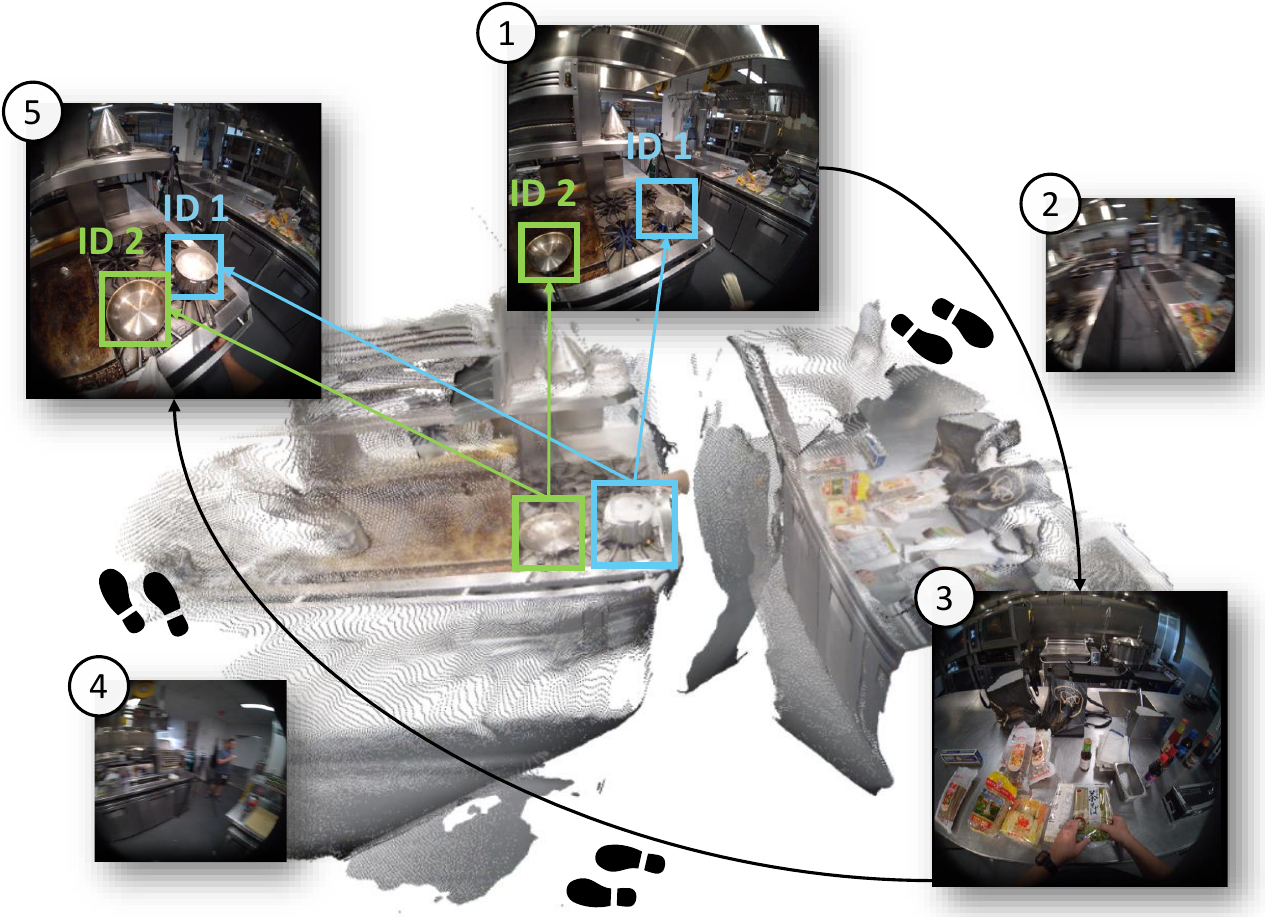}\\
    \caption{An illustrative example of Ego3DT. It showcases robust 3D object tracking across ego-centric video frames (from Frame 1 to Frame 5). The 3D field maintains consistent object information, ensuring the tracking ID remains unchanged. This delivers reliable tracking results in dynamic video scenarios, as shown by the persistent tracking of \textbf{ID 1} and \textbf{ID 2} across different viewpoints.
    }
    \Description{An illustrative example of Ego3DT.}
    \label{fig:fig_teaser}
\end{figure}

\section{Introduction}
Ego-centric, or first-person, computer vision addresses the perceptual challenges an embodied AI encounters in real-world situations. This area has garnered significant interest due to its relevance in various applications, including robotics~\cite{savva2019habitat,duan2022survey}, embodied agents~\cite{wang2023embodiedscan, zhao2023see, zhao2024hierarchical, zhao2024we,zhao2024steve}, and mixed reality~\cite{ego-exo,grauman2022ego4d,grauman2023ego,deng2024citycraft}. One of the core tasks in this field is multi-object tracking (MOT), which involves object detection, re-identifying objects in the environment, and predicting the future state of the surroundings.

Despite significant advancements in MOT~\cite{hao2024divotrack,wang2019exploit,wang2021track,wang2022split}, applying these methods to ego-centric videos remains underexplored. This gap is largely attributed to the absence of comprehensive ego-centric tracking datasets, essential for training and evaluating tracking algorithms~\cite{fan2019lasot}. Although the research community has introduced several popular tracking datasets such as OTB~\cite{wu2013online}, TrackingNet~\cite{muller2018trackingnet}, GOT-10k~\cite{huang2019got}, and LaSOT~\cite{fan2019lasot}, the existing trackers are not proven high performance for ego-centric videos, and the benchmarks lack such videos and comprehensive annotations for all object tracklets. The lack of benchmarks for verifying SOTA trackers' performance underscores the urgent need for a dedicated ego-centric tracking dataset, particularly one that can support the unique requirements of ego-centric applications.

Differing from traditional third-person videos, ego-centric videos often capture a wide range of activities, objects, and locations without a specific focus. Large head movements from the camera wearer frequently cause objects to exit and re-enter the field of view, and objects manipulated by hands may undergo frequent occlusions, along with rapid changes in scale, pose, and even state or appearance~\cite{shan2020understanding}. These unique aspects make object tracking significantly more demanding than in scenarios typically presented in existing datasets, highlighting a critical gap in current evaluation methodologies. Traditional MOT tasks~\cite{zhang2022bytetrack}, when applied to ego-centric videos~\cite{tang2024egotracks}, often result in poor tracking accuracy and robustness.

To address variations in ego-centric videos, we propose \textbf{Ego3DT} which uses a 3D field representation for more robust tracking. As shown in Figure~\ref{fig:fig_teaser}, the 3D field captures the spatial layout and the temporal dynamics of objects within the scene, making it exceptionally suitable for the complexities of ego-centric views. This concept involves maintaining a dynamic 3D scene to enhance perceptual tasks~\cite{wang2023embodiedscan, wang2023dust3r}. 3D perception can improve task robustness by ensuring stable object properties and relationships throughout the scene. By maintaining a dynamic 3D field, our approach preserves stable relationships and properties of 3D objects, significantly enhancing performance. Moreover, our few-shot method employs training-free, plug-and-play modules, distinguishing it from conventional approaches. We summarize our contributions as follows:
\begin{itemize}
\item We propose a method for constructing a 3D scene from an ego-centric video and achieving open-vocabulary object tracking, which requires only RGB videos as input and is a zero-shot approach.

\item We implement object 3D position matching through a dynamic cross-window matching method, thereby alleviating the instability caused by relying solely on 2D image tracking.

\item Our method achieves state-of-the-art performance on the open-vocabulary multi-object tracking in ego-centric videos, with $1.04 \times$ - $2.90\times$ in HOTA.
\end{itemize}
\section{Related Work}

\subsection{Open-Vocabulary Detection}
Open-vocabulary~(OV) detection~\cite{OVD} has emerged as a novel approach to modern object detection, which aims to identify objects beyond the predefined categories. Early studies~\cite{ViLD} followed the standard OV Detection setting~\cite{OVD} by training detectors on the base classes and evaluating the novel or unknown classes. However, this open-vocabulary setting, while capable of evaluating the detectors' ability to detect and recognize novel objects, is still limited to open scenarios and lacks generalization ability to other domains due to training on a limited dataset and vocabulary. Inspired by vision-language pre-training~\cite{ALIGN}, recent works~\cite{RegionCLIP,DetPro,BARON} formulate open-vocabulary object detection as image-text matching and exploit large-scale image-text data to increase the vocabulary at scale. GLIP~\cite{GLIP} presents a pre-training framework for open-vocabulary detection based on phrase grounding and evaluates in a zero-shot setting. Grounding DINO~\cite{GroundingDINO} incorporates the grounded pre-training into detection transformers~\cite{DINO} with cross-modality fusions. Several methods~\cite{GLIPv2, DetCLIP, wu2023general} unify detection datasets and image-text datasets through region-text matching and pre-train detectors with large-scale image-text pairs, achieving promising performance and generalization. 
% However, these methods often use heavy detectors, leading to high computational demands and deployment challenges. Utilizing the YOLO framework with an effective pretraining strategy, some works~\cite{ZSD_YOLO,cheng2024yoloworld} enhance open-vocabulary performance and generalization. GLEE~\cite{wu2023general} excels in recognizing and tracking objects across both images and videos.

\subsection{Ego-centric Tracking}
Over the last few decades, the introduction of numerous ego-centric video datasets~\cite{Damen2018EPICKITCHENS,grauman2022ego4d}, has presented a wide range of fascinating challenges. Although many methodologies utilize tracking to address these challenges~\cite{grauman2022ego4d,huang2024exploring}, it's notable that only a few studies have focused solely on the crucial issue of tracking. The works by Dunnhofer et al.~\cite{dunnhofer2021first, dunnhofer2023visual} address the specific challenges associated with ego-centric object tracking and represent the research most closely related to our own. However, a significant distinction exists in the scale of the dataset they utilized, which comprises 150 tracks designed purely for assessment purposes.
In ego-centric video comprehension, Ego4D~\cite{grauman2022ego4d}, EPIC-KITCHENS VISOR~\cite{darkhalil2022epic} and EgoTracks~\cite{tang2024egotracks} are critical to our work. Ego4D stands out for its extensive compilation of ego-centric videos captured in natural settings and introduces numerous innovative tasks, including object tracking. Concurrently introduced, VISOR focuses on annotating brief videos (averaging 12 seconds in length) from EPIC-KITCHENS~\cite{Damen2018EPICKITCHENS} with instance segmentation masks, illustrating the dynamic and detailed nature of this field.

\subsection{Ego-centric 3D Understanding}
The study of 3D object detection has made considerable advancements through the utilization of images~\cite{rukhovich2022imvoxelnet,sparf,liu2024monotakd,deng2023citygen,zhang2019eye}, point clouds~\cite{pix4point,Li2022sctn}, and videos~\cite{caesar2020nuscenes,chai2023stablevideo,song2023moviechat,song2024moviechat+}. To convert 2D images into 3D scenes, researchers have extensively employed Structure from Motion (SfM) techniques~\cite{ozyecsil2017survey}. These techniques are divided into geometric-based methods~\cite{schoenberger2016sfm,labbe2019rtab,mur2015orb}, which rely on multiview geometry; learning-based methods~\cite{zhou2017unsupervised,vijayanarasimhan2017sfm,kendall2015posenet}, which utilize deep neural networks; and hybrid SfM approaches~\cite{teed2018deepv2d,teed2021droid}, which integrate both strategies. SfM has been adapted for extensive videos in dynamic settings~\cite{zhao2022particlesfm} and casual videos capturing everyday life~\cite{zhang2022structure,liu2022depth}.
However, existing methods struggle to be effective due to the dynamic views and motion blur of ego-centric videos. Numerous studies explore the reconstruction of 3D human poses from ego-centric footage~\cite{rhodin2016egocap,wang2021estimating,li2023ego, zhang2022egobody,dai2022hsc4d}. Ego-HPE~\cite{park2023domain} tackled the challenges of ego-centric 3D human pose estimation with their domain-guided spatiotemporal transformer model. There are also some efforts including the investigation into ego-centric indoor localization using the Manhattan world assumption for room layouts~\cite{chen2022egocentric}. Existing 3D scene generation methods aim to generate unknown 3D scenes from 2D layouts or user definitions.
% Additionally, some work~\cite{nagarajan2022egocentric} proposed a method that correlates camera positions with video data to anticipate human-centric scene contexts.
Our approach focuses on 3D tracking via dynamic matching in the 3D field. It is a zero-shot, RGB-only approach for open-vocabulary object tracking by 3D constructing from the ego-centric video.

\begin{figure*}[t]
% \vspace{-30pt}
% \setlength{\tabcolsep}{1.5pt}
\centering
    \includegraphics[width=0.95\linewidth]{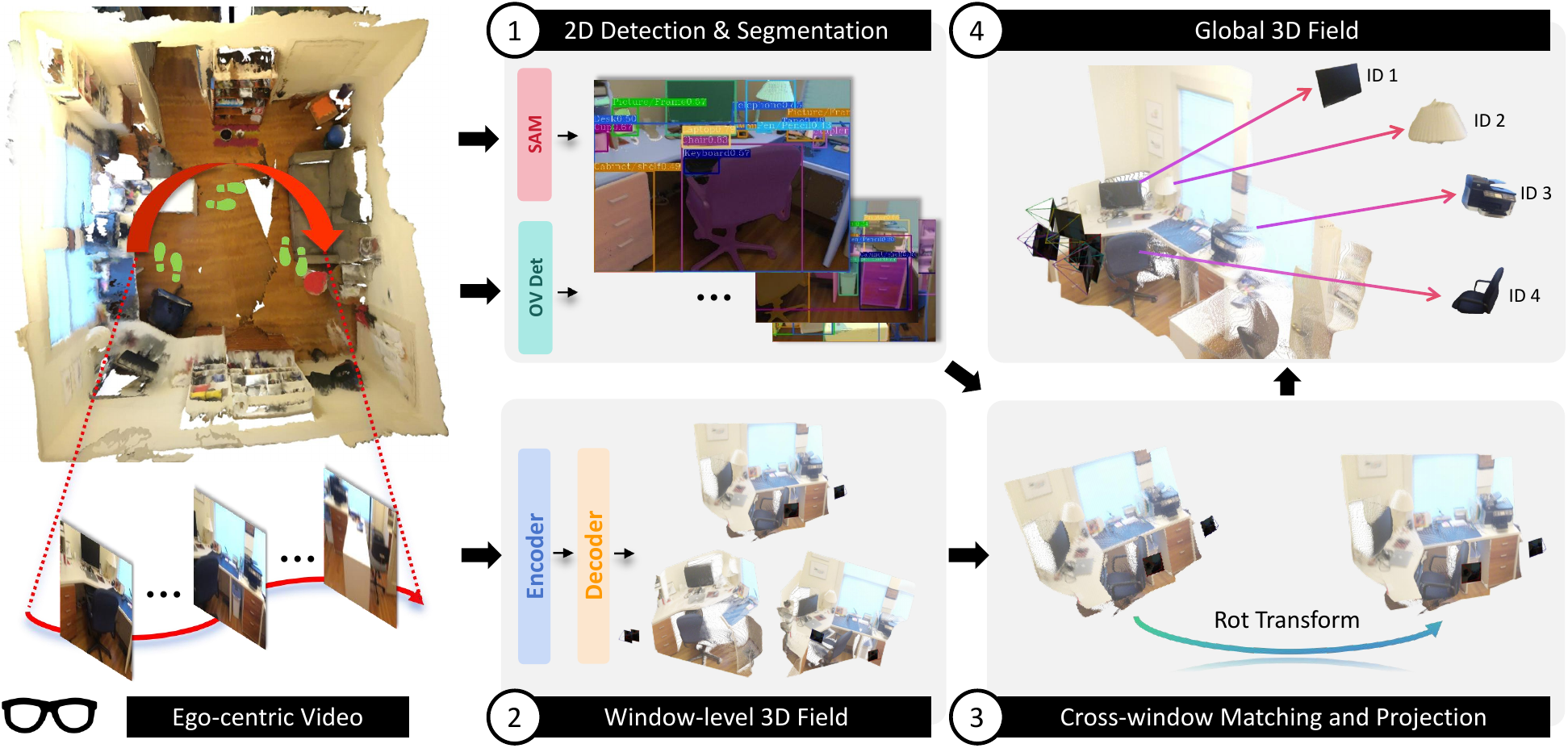}\\
    % \vspace{-5pt}
    \caption{\textbf{Ego3DT framework. (1) 2D Detection \& Segmentation: Ego-centric video frames undergo object detection and segmentation using SAM to segment object points and an OV detector to identify objects. (2) Window-level 3D Field: The encoder-decoder structure processes the segmented frames to construct a window-level 3D field. (3) Cross-window Matching and Projection: Subsequent windows are aligned using rotational transforms to maintain object consistency across frames. (4) Global 3D Field: The cumulative data from all windows is integrated to form a global 3D field, with each object assigned a unique ID, facilitating precise object tracking throughout the video sequence.} 
    }
    \Description{The framework of Ego3DT.}
    \label{fig:framework}
\end{figure*}

\section{Method}
\subsection{Overview}

As shown in Figure~\ref{fig:framework}, Ego3DT is a purely vision-based open-vocabulary 3D object tracking method $\mathcal{F}$ to achieve tracking results $Y$ from RGB ego videos $X$ containing frames from $I_1$ to $I_N$. The open-vocabulary object tracking results $Y$ can be obtained as follows,
\begin{equation}
Y = \mathcal{F}(X), \quad X = [I_1, I_2, ..., I_N], 
\label{eqn:main_function}
\end{equation}
where $Y = \{O_i\}_{i\leq N}$ is the 3D object tracking output of the video with $N$ frames, $O_i = [(x_j, y_j, z_j, \mathbf{ID}_j)]_{j\leq K}$ is a matrix containing 3D coordinates of tracked objects in each frame with identification $\mathbf{ID}$, and $K$ is the total number of tracked objects.

First, we conduct object detection $\mathbf{Det}$ on videos $X$, and semantic segmentation $\mathbf{Seg}$ based on detection output $O^{Det}_{2D}$ as prompts:
\begin{equation}
O^{Seg}_{2D} = \mathbf{Seg}(O^{Det}_{2D}), \quad O^{Det}_{2D} = \mathbf{Det}(X), 
\label{eqn:seg_det}
\end{equation} 
where $O^{Seg}_{2D}$ and $O^{Det}_{2D}$ are the semantic segmentation and detection output respectively.
 
Then, we utilize a 3D estimation model $\mathcal{G}$ to map segmentation coordinates from 2D space $O^{Seg}_{2D}$ to 3D space $O_{3D} \in \mathbb{R}^{K \times N \times 3}$:
\begin{equation}
O_{3D} = \mathcal{G}(X, O^{Seg}_{2D}), 
\label{eqn:estimation}
\end{equation} 
where 
% $X$ its corresponding RGB video $X$ of frame number $N$ with pixel number of objects $K$, 
$O_{3D}$ forms a one-to-one mapping between image pixels and 3D scene points, \ie, $O_{2D} \leftrightarrow O_{3D}$, for all object coordinates $(x,y) \in \{1\ldots K\}\times\{1\ldots N\}$. 

Finally, Ego3DT involves matching the 3D positions of objects using a hierarchical method, avoiding the instability issues that can arise from relying solely on 2D image tracking:
\begin{equation}
Y = \mathcal{M}(O_{3D}) = \mathbf{PointMatch}(\mathcal{A}(O_{3D})),
\label{eqn:match}
\end{equation} 
where the matching module $\mathcal{M}$ compares all the 3D points from frame to frame for precise object tracking $Y$ with identification $\mathbf{ID}$, and $\mathcal{A}$ is a 3D scene registration method aligning adjacent points. We use the additional Hungarian process to initialize matching $\mathbf{ID}$.

% \vspace{-6pt}
\subsection{2D Segmentation and Open-Vocab Detection}

The foundational step in our method involves the precise identification and segmentation~\cite{hao2021weakly} of objects within each frame of an ego-centric video. As shown in Equation~(\ref{eqn:seg_det}), this process is bifurcated into two pivotal operations: 2D open-vocabulary (OV) Detection $\mathbf{Det}$ and 2D segmentation $\mathbf{Seg}$, applied sequentially to the raw video frames to ensure a comprehensive understanding of the scene.

To achieve accurate object detection within our framework, we leverage the capabilities of the pre-trained GLEE~\cite{wu2023general} in the experiment. Our efficient object detection model can identify a wide range of objects in 2D space across video frames, even those not explicitly labeled in the training data. We obtain precise 2D bounding boxes for all detectable objects by processing each frame through the model, setting the stage for subsequent segmentation.

Following the detection phase, the identified objects are further processed through SAM~\cite{kirillov2023segment}, a segmentation foundation model designed to delineate the precise boundaries of objects within an image. The bounding boxes obtained from GLEE~\cite{wu2023general} serve as prompts for SAM~\cite{kirillov2023segment}, enabling it to focus on specific regions of interest within the frame. This approach generates detailed segmentation maps for each object, including shape and location.

\subsection{Window-level 3D Fields}

We maintain window-level 3D fields with a 3D estimation model called $\mathcal{G}$, a pre-trained DUSt3R~\cite{wang2023dust3r}. This dual-branch system consists of image encoders, decoders, and regression heads. The image encoders are designed to extract detailed feature maps from segmented 2D object points, which are inputs derived from the preceding object detection and segmentation phases. The decoders then process these feature maps, focusing on extracting spatial relationships and depth cues from the encoded data. As shown in Equation~(\ref{eqn:estimation}), the 3D estimation model $\mathcal{G}$ processes segmented 2D object points, transforming them into their 3D counterparts through the pretrained DUSt3R~\cite{wang2023dust3r}. This process is predicated on accurately detecting and segmenting objects within the 2D space, followed by their elevation into the 3D domain. 

\paragraph{From 2D Segmentation to 3D Localization.} As shown in Figure~\ref{fig:framework}, the image encoders are designed to extract detailed feature maps from segmented 2D object points, inputs derived from the preceding object detection and segmentation phases. The decoders then process these feature maps, focusing on extracting spatial relationships and depth cues from the encoded data.

\paragraph{Integration and Alignment of 3D Data.} The output from $\mathcal{G}$ consists of accurate 3D coordinates inherently aligned with the original RGB video frames. This alignment is critical as it ensures that each 3D point precisely represents its corresponding 2D point and is correctly positioned within the global context of the video sequence. This meticulous alignment facilitates the seamless integration of 2D and 3D data, enhancing the robustness and accuracy of the subsequent object-tracking processes.

% Keeping window-level 3D fields in our framework advances the field of 3D object tracking and sets a new standard for the accuracy and efficiency of converting 2D video data into actionable 3D information. The rigorous processing and alignment of data ensure that our model is highly effective in the challenging environment of ego-centric videos, paving the way for innovative applications in multiple domains.

\subsection{Cross-window Matching and Projection}

The Matching Module $\mathcal{M}$ is a crucial component of the Ego3DT framework for tracking 3D objects across video sequences. As shown in Equation~(\ref{eqn:match}), the Matching Module $\mathcal{M}$ consists of point-matching algorithms and a sliding window mechanism to ensure accurate and robust object tracking, even in occlusion or rapid movements.
To minimize errors in point matching, we retain mutual correspondences between two images. This is achieved by performing the KDTree search~\cite{wang2023dust3r} in the 3D pointmap space.

\paragraph{Sliding Window Mechanism.}
We adapt the sliding window mechanism in the matching module, defined by the window size ${W}$, ensuring an overlap size ${T}$ to maintain temporal continuity between frames. This design choice allows for the efficient processing of video frames by dividing the extensive task of 3D object tracking into manageable segments, each containing ${W}$ frames. The step distance $S = {W} - {T}$ dictates the window's movement across the video sequence, ensuring that every frame is analyzed while optimizing computational resources.

\paragraph{Initial Object Tracking.}
The process begins by establishing a baseline of object tracking within the first window. For each frame $i$, up to the window size ${W}$, the 3D coordinates of detected objects $O_{3D}^i = \{(x_j, y_j, z_j)\}_{j \leq K}$ are determined, where $K$ represents the number of objects detected within a frame. Utilizing KDTree distance calculations between every two consecutive frames, we employ the Hungarian algorithm to match objects based on their spatial proximity, thus assigning a unique Identification Number $\mathbf{ID}$ to each object. The result, $Y_0$, comprising tracked objects with their respective $\mathbf{ID}$s within the first window, is stored in a buffer $\mathcal{B}$ for subsequent processing.

\paragraph{Dynamic matching across windows.}
The module employs a hierarchical object-tracking approach as shown in the Algorithm~\ref{alg:match}. As the window slides by step $S$, each new set of frames is processed based on the previous window's data. Specifically, we employ a 3D scene registration method $\mathcal{A}$, an optimized homography process to align the 3D points of objects between the current and previous windows, thus $O_{3D}^t = \mathcal{A}(O_{3D}^{t-1}, O_{3D}^t)$ to keep the current windows $O_{3D}^t$ into the same space of the previous $O_{3D}^{t-1}$. The homography process is shown as follows:
\begin{equation}
    O_{3D}^{t-1} = H^{t \rightarrow t-1} O_{3D}^t,
\end{equation}
where $H^t$ is the homography matrix between the current points $O_{3D}^{t}$ and the previous points $O_{3D}^{t-1}$ in the sliding windows, the ground points of all current frames are unified into the previous space. 
To further refine the alignment process, $\mathcal{M}$ employs an optimization strategy that minimizes the Euclidean distance between matched points across the homography transformations:
\begin{equation}
    H^{t}_{*} = \argmin_{\mathbf{T}} \frac{1}{A} \sum_{t=1}^T ||O_{3D}^{t-1} - H^t O_{3D}^t||_2, 
\end{equation}
where $A$ is the total number of matching points, $H^t$ is a $4 \times 4$ matrix, $T$ is the overlap size. During initialization, all parameters are random numbers in the $(0, 1)$ range.

By recalculating KDTree distances for the newly aligned 3D points and based on the applied Hungarian algorithm, $\mathbf{PointMatch}$ matches pixels of objects from frame to frame. Each object in the current window is then assigned the $\mathbf{ID}$ of its closest match from the previous window, thus extending the tracking sequence. This process is repeated for each window throughout the video, culminating in comprehensively tracking all objects across the sequence.

The Matching Module $\mathcal{M}$ of Ego3DT achieves high precision and robustness in 3D object tracking through these sophisticated algorithms and mechanisms. It provides a global 3D field as shown in Figure~\ref{fig:case}. This innovative approach ensures that Ego3DT can effectively handle the complexities of ego-centric video analysis, paving the way for advancements in interactive and immersive technologies. We summarize the matching process in Algorithm~\ref{alg:match}, which outlines the step-by-step procedures for achieving accurate and reliable tracking results. 

\begin{algorithm}[t]
\caption{Cross-window Matching Process $\mathcal{M}$}
\label{alg:match}
\begin{algorithmic}[1]
\STATE \textbf{Input:} Video frames $X = \{I_i\}_{i=1}^N$, Initial 3D coordinates $O_{3D}^1$, Window size ${W}$, Overlap size ${T}$
\STATE \textbf{Output:} Tracked objects ${Y}$ with IDs
\STATE \textbf{Initialize:} Buffer $\mathcal{B} \leftarrow \emptyset$, Detector $\mathbf{Det}$, Segmenter $\mathbf{Seg}$, 3D Estimator $\mathcal{G}$
\STATE $Y_0 \leftarrow Hungarian(\mathbf{PointMatch}(O_{3D}^1))$ 
\STATE Add $Y_0$ to $\mathcal{B}$ // Save to memory.
\STATE // Cross-window matching in the overlap
\FOR{$t = 1$ to $T$ } 
    \STATE $O_{3D}^t \leftarrow \mathcal{G}(X, \mathbf{Seg}(\mathbf{Det}(I_t)))$
    \STATE Align 3D scenes: $O_{3D}^t \leftarrow \mathcal{A}(O_{3D}^{t-1}, O_{3D}^t)$
\ENDFOR
\FOR{$t = 1$ to $W$ }
    \STATE $Y_t \leftarrow \mathbf{PointMatch}(O_{3D}^{t-1}, O_{3D}^t)$ // Matching 3D points
    \STATE Add $Y_t$ with IDs to $\mathcal{B}$ // Save to memory.
\ENDFOR
\STATE Convert buffer $\mathcal{B}$ to the output space $Y$
\STATE \textbf{return} $Y$
\end{algorithmic}
\end{algorithm}

\begin{table*}[t]
    \centering
    \tiny
    \caption{\textbf{Comparison of Open Vocabulary MOT performance.} 2D box and 3D point refer to association to 2D box and 3D point. ``$f$'' stands for feature association.}
    \resizebox{1\linewidth}{!}{
    \begin{tabular}{c | c | c | cccccc}
    \toprule
        Tracker & Detector & Association  & HOTA~($\uparrow$) & IDF1~($\uparrow$) & DetA ($\uparrow$) & MT ($\uparrow$) & ML ($\downarrow$) & Frag ($\downarrow$) \\
    \midrule\midrule
        \multirow{2}{*}{ByteTrack~\cite{zhang2022bytetrack}} 
        & YOLO-World~\cite{cheng2024yoloworld} & 2D box &  19.14 & 18.77 & 17.11 & 23 & 78 & 775\\
        & GLEE~\cite{wu2023general}  & 2D box & 29.58 & \textbf{31.28} & 29.10 & \textbf{30} & 73 & 1217\\
    \midrule
         \multirow{2}{*}{DeepSort~\cite{wojke2017simple}} 
        & YOLO-World~\cite{cheng2024yoloworld} & 2D box + $f$ & 10.63 & 9.63 & 11.15 & 9	& 106 & 637\\ 
         & GLEE~\cite{wu2023general} & 2D box + $f$ & 15.91 & 15.79 & 18.00 & 9 & 90 & 710\\
    \midrule
        OVTrack~\cite{li2023ovtrack} & OVTrack~\cite{li2023ovtrack} & 2D box + $f$ & 15.40 & 15.15 & 12.90 & 6 & 123 & 816\\
        TET~\cite{trackeverything} & TET~\cite{trackeverything} &  2D box + $f$  & 13.94 & 13.34 & 11.41 &  5	&  134	&  583\\ 
    \midrule\midrule
        \multirow{4}{*}{\textbf{Ego3DT~(Ours)}} & OVTrack~\cite{li2023ovtrack} & \multirow{4}{*}{3D point}  & 13.44 & 12.90 & 13.79 & 5 & 138 & 512\\
         & TET~\cite{trackeverything} & & 12.40 & 11.62 & 13.24 & 5 & 134	& \textbf{463}\\ 
         & YOLO-World~\cite{cheng2024yoloworld} & & 16.28 & 15.28 & 19.43 & 14 & 78 & 1196 \\ 
         & \textbf{GLEE}~\cite{wu2023general}  & & \textbf{30.83} & 29.71 & \textbf{47.91} & 24 & \textbf{49} & 1217\\
    \bottomrule
    \end{tabular}
    }
    \label{tab:MOT}
\end{table*}
\section{Experiment}

This section evaluates the Ego3DT framework for 3D object tracking in ego-centric videos using the Ego3DT Benchmark. We form two datasets: Ego3DT-daily and Ego3DT-indoor, and advance metrics to evaluate tracking accuracy. We test the state-of-the-art detectors and compare their performance to baseline models, demonstrating the efficacy and robustness of the Ego3DT in handling the unique challenges of ego-centric video analysis. Through rigorous testing and validation, this section illustrates the robustness, precision, and scalability of the Ego3DT framework.

\subsection{Ego3DT Benchmark}
Since there is no existing multi-object tracking benchmark based on ego-centric videos, we build a new benchmark called Ego3DT Benchmark to evaluate the performance of our model. 
\subsubsection{Datasets Description}
We collected and re-annotated two datasets, Ego3DT-daily and Ego3DT-indoor, from Ego4D~\cite{grauman2022ego4d} and EmbodiedScan~\cite{wang2023embodiedscan}. These datasets include 2D detection boxes and daily object trajectories in indoor and outdoor scenes.

\paragraph{Ego3DT-daily} includes six indoor and outdoor scenes from Ego4D videos~\cite{grauman2022ego4d}. Each video has 500 consecutive frames sampled at 10 FPS. There are two outdoor scenes and four indoor scenes. The video collection locations include supermarkets, gardens, corridors, and kitchens. These ego-centric videos feature noticeable shaking and diverse object changes.

\paragraph{Ego3DT-indoor} includes data from five indoor scenes. Based on the Embodied Scan dataset, we collected ego-centric videos following predefined camera trajectories. We collected about 100 frames per video at 3 FPS from five scenes.

% \begin{table*}[!t]
% \caption{\textbf{Ablation studies.} }
% \centering
% \resizebox{0.7\linewidth}{!}{
% \begin{tabular}{l|cc|cc|cc}
% \toprule
% \multirow{2}{*}{Matrix} & \multicolumn{2}{c|}{Detector} & \multicolumn{2}{c|}{Segmentor} & \multicolumn{2}{c}{Memory}\\
% \cmidrule(lr){2-3} \cmidrule(lr){4-5} \cmidrule(lr){6-7}
% & GLEE~\cite{wu2023general} & Ground Truth & SAM~\cite{kirillov2023segment} & GLEE~\cite{wu2023general} & Full & 30 & 0  \\
% \midrule
% HOTA~($\uparrow$) & & & & & & \\ 
% \midrule
% IDF1~($\uparrow$) & & & &  &  & \\ 
% \midrule
% DetA~($\uparrow$) & & & & & & \\ 
% \bottomrule
% \end{tabular}
% }
% \label{tab:ablation}
% \end{table*}

\begin{table*}[t]
\centering
\tiny
\caption{Ablation study with different detectors and memory mechanisms of varying strengths.}
\label{tab:Ablation}
\resizebox{0.9\linewidth}{!}{
\begin{tabular}{c|c|cccccc}
\toprule
     \multicolumn{2}{c|}{Setting} & HOTA~($\uparrow$) & IDF1~($\uparrow$) & DetA ($\uparrow$) & MT ($\uparrow$) & ML ($\downarrow$) & Frag ($\downarrow$) \\ \midrule%\midrule
     \multirow{2}*{Detector}  
        & \multirow{1}*{YOLO-World}~\cite{cheng2024yoloworld}
             & 16.28 & 15.28 & 19.43 & 14 & 78 & \textbf{1196}  \\
        & \multirow{1}*{GLEE}~\cite{wu2023general}
             & \textbf{30.83} & \textbf{29.71} & \textbf{47.91} & \textbf{24} & \textbf{49} & 1217\\
    \midrule %\midrule
     \multirow{3}*{Memory} 
        & \multirow{1}*{w/o Memory}
            & 29.13	& 28.68	& 44.56	& 21 & \textbf{49} & \textbf{1216}\\
        & \multirow{1}*{30 Frames}
            & \textbf{30.83}	& \textbf{29.71}	& \textbf{47.91}	& \textbf{24} & \textbf{49} & 1217\\
        & \multirow{1}*{Full Frames}
            & 27.60 & 28.54 & 38.60 & 18 & 109 & 1241\\
\bottomrule 
\end{tabular}
}
\end{table*}

\subsubsection{Annotation and Metrics}

Our annotation pipeline is semi-automatic. We annotated the same objects with detection boxes and a global ID in a single video. For the Ego3DT-daily dataset, we first used the existing open vocabulary detector GLEE to extract object detection boxes to save annotation time. We then calibrated and aligned each object's detection boxes and IDs frame by frame. For objects that disappeared and then reappeared, we assigned them a consistent global ID. For the Ego3DT-indoor dataset, since Embodied Scan provides 3D detection boxes for each object, we projected the 3D detection boxes onto the current frame based on the camera's pose in each frame, thus determining the object's 2D detection boxes and global ID.

We evaluate the performance of our method using HOTA~\cite{luiten2021hota} and the MOT Challenge~\cite{dendorfer2021motchallenge} evaluation metrics, including IDF1, MT, ML, Frag~\etc ~IDF1 assesses the consistency of IDs and places more emphasis on association performance. HOTA explicitly balances the accuracy of detection, association, and localization.

\subsection{Experiment Setups}

We conduct experiments on Ego3DT using different detectors for open vocabulary detection, namely GLEE~\cite{wu2023general} via GLEE-Plus backbone Swin-L and YOLO-World~\cite{cheng2024yoloworld} via YOLO-Worldv2-X. We also use SAM~\cite{kirillov2023segment} with ViT-H backbone for open vocabulary segmentation. Then, we utilize the 3D estimation model via DUSt3R~\cite{wang2023dust3r} with DPT Head, ViT-L Encoder, and ViT-B Decoder. Note that our experiments are conducted using only a single RTX3090-24G.

\subsection{Baselines}
We evaluate the Ego3DT framework against established baselines: ByteTrack~\cite{zhang2022bytetrack}, DeepSort~\cite{wojke2017simple}, OVTrack~\cite{li2023ovtrack}, and TET~\cite{trackeverything}, each offering unique strengths in multi-object tracking (MOT) and providing a comprehensive comparison of our model.

\textit{ByteTrack}~\cite{zhang2022bytetrack} is a powerful multi-object tracking (MOT) method designed to associate each detection box, regardless of the score, to improve tracking consistency, especially in cases with occluded objects. It stands out due to its simplicity, efficiency, and robustness against occlusions and low-confidence detections. ByteTrack has been successfully applied to different tracking benchmarks, confirming its versatility and strength as a baseline model.

\textit{DeepSort}~\cite{wojke2017simple} is an effective MOT method in videos, enabling accurate identity retention over time, particularly in scenarios where objects are frequently occluded. This system is a go-to choice for practitioners seeking a balance between performance and computational efficiency.

\textit{OVTrack}~\cite{li2023ovtrack} is an open-vocabulary MOT method, utilizing vision-language models for classification and association, applying knowledge distillation and data hallucination techniques for feature learning. The approach aims to be highly data-efficient and is tailored for large-scale tracking.

\textit{TET}~\cite{trackeverything} is a large-scale MOT method. It critically examines the limitations of current MOT metrics and methods, which often assume near-perfect classification performance, a presumption rarely met in practice. TET performs associations using Class Exemplar Matching, showing notable improvements in challenging tracking.

\begin{figure*}[ht]
\centering
    \includegraphics[width=\linewidth]{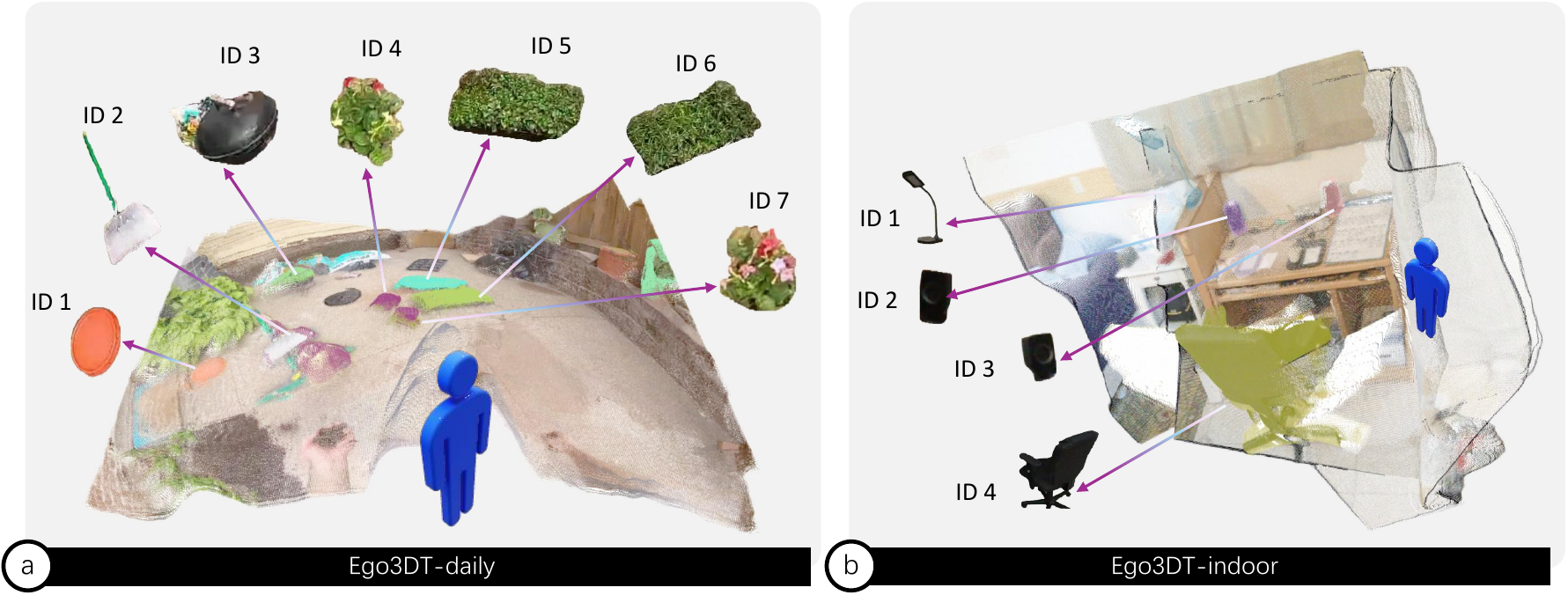}\\
    \caption{Qualitative results of the 3D tracking field in Ego3DT: a) For the Ego3DT-daily dataset, diverse outdoor objects (IDs 1-7) are successfully tracked within the environment, showing the model's capability to handle varying object types and outdoor conditions. b) In the Ego3DT-indoor dataset, common indoor objects (IDs 1-4) are tracked with high fidelity in a typical room setup, demonstrating the precision of the 3D tracking across different indoor scenes.}
    \Description{Qualitative results of the 3D tracking field in Ego3DT.}
    \label{fig:case}
\end{figure*}

\subsection{Evaluation Results}

As shown in Table~\ref{tab:MOT}, we have evaluated the open-vocabulary multi-object tracking performance using a comprehensive range of metrics from the MOT Challenge and HOTA. Ego3DT greatly outperforms well-established baselines with a unique approach to 3D point association. It has been assessed on additional performance indicators, thus enhancing the breadth of our evaluation.

Notably, the Ego3DT framework with the GLEE detector~\cite{wu2023general} achieves the highest HOTA score of 30.83 among all evaluated trackers, indicative of a well-balanced detection and association accuracy.
It excels in DetA (Detection Accuracy) with a leading score of 47.91, demonstrating our framework's exceptional capability in precise object detection. Ego3DT improves with
better detector performance. Note that DetA is not the same across different methods, even if the same detector is used. This is because different methods adopt different association and post-processing strategies that may affect the detection results.
Furthermore, Ego3DT maintains a competitive edge with lots of Mostly Tracked (MT) targets and the fewest Mostly Lost (ML) targets among the automatic tracking methods, with respective scores of 24 and 49, highlighting the framework's robustness in persistent object tracking over time. The TET detector~\cite{trackeverything} produces the highest number of Fragmentations (Frag), indicating that our tracking is accurate and the object identity is stable compared to the object trajectories in the ground truth.

These expanded metrics provide a holistic view of our framework's performance, affirming its strengths in maintaining object identities (as evidenced by its IDF1 score of 29.71) and effectively tracking objects throughout the video sequence. Despite the high Frag count, the Ego3DT framework excels in key metrics, proving it to be a robust solution for the MOT in ego-centric videos.

\subsection{Ablation Study}
To refine the Ego3DT framework, we conduct a comprehensive ablation study to discern the individual contributions of detector quality and memory mechanisms to the framework's overall performance. The experiments are carefully designed to isolate the impact of these components, providing insights into their respective significance and interplay. As shown in Table~\ref{tab:Ablation}, a high-quality detector profoundly influences the framework's performance, and memory mechanisms play a nuanced role in achieving state-of-the-art tracking performance in open vocabulary MOT scenarios.

\paragraph{Accurate Detector is Pivotal.} We select the GLEE detector~\cite{wu2023general} as the high-quality pre-trained detector, yielding a HOTA score of 30.83, which robustly indicates superior detection and ID association. Comparing YOLO-world and GLEE, Table~\ref{tab:MOT} shows GLEE outperforms YOLO-world on DetA by 11.99\%, 6.85\%, and 28.48\% in ByteTrack, DeepSort, and Ego3DT, respectively. This confirms that a proficient detector can substantially boost the tracking quality.

\paragraph{Appropriate Memory Mechanism is Critical.} It reflects on the balance between memory usage and tracking performance. Notably, using a 30-frame memory mechanism offers the best performance across all metrics. This optimized setting achieves a HOTA of 30.83 and an IDF1 of 29.71, underscoring the effectiveness of a limited temporal memory that captures the immediate past to maintain context without being burdened by the noise of distant frames. On the other hand, the absence of a memory mechanism and the use of full-frame memory result in reduced performance, demonstrating the importance of a focused temporal window for accurate tracking. This suggests that an excessive memory span can dilute the relevancy of information, leading to higher fragmentation and decreased detection accuracy. The results highlight the trade-off between the size of memory and the tracking accuracy, suggesting that moderate memory size is instrumental in improving the consistency and precision of object tracking in ego-centric videos.

\begin{figure*}[ht]
\centering
    \includegraphics[width=\linewidth]{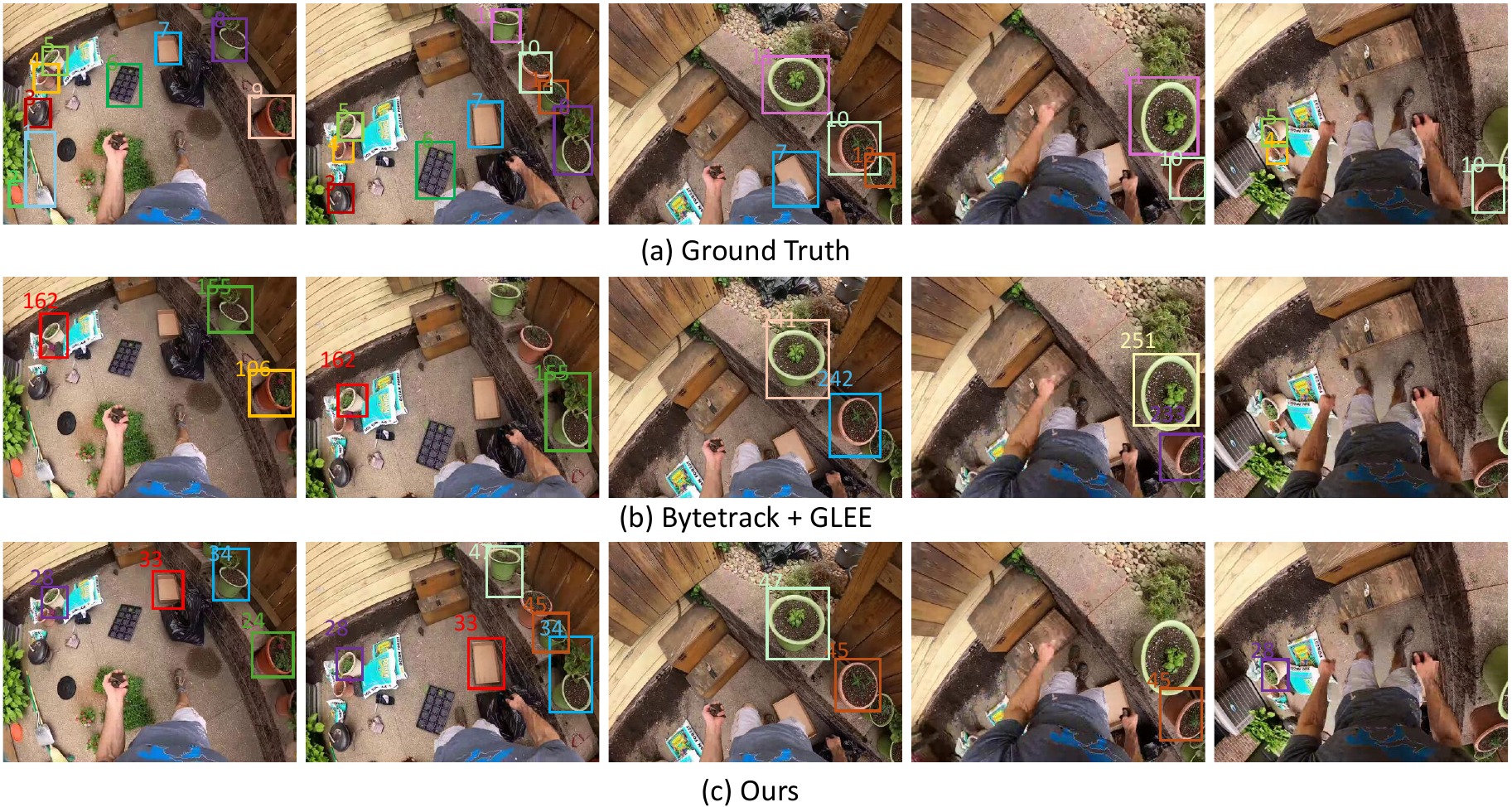}\\
    \caption{Qualitative results of 2D tracking comparison: a) Ground Truth sequence showing accurate object detection and consistent ID assignment over time. b) ByteTrack with GLEE detection demonstrating object tracking and identification, with occasional ID inconsistencies and missed detections. c) Our Ego3DT approach maintains stable object identification, accurately captures dynamic objects, and excels in consistent ID assignment, especially in motion-rich ego-centric views. From left to right represents the tracking results of each method over time.}
    \Description{Qualitative results of 2D tracking comparison.}
    \label{fig:case_track}
\end{figure*}

\subsection{Qualitative Results}

Our Ego3DT framework exhibits significant advancements in 3D reconstruction and 2D tracking, showcasing robust performance even under challenging first-person motion scenarios. We provide a qualitative analysis of these two core aspects to highlight the efficacy and improvements over existing methodologies.

\subsubsection{Qualitive Results on 3D Reconstruction.}
The performance of Ego3DT on our dataset demonstrates its ability to handle complex 3D environments. As shown in Figure ~\ref{fig:case}, we present two different datasets where Ego3DT accurately tracks each object in both indoor and outdoor scenes.

\paragraph{Outdoor Tracking in Ego3DT-daily.} The Ego3DT-daily dataset, representing an array of outdoor settings, challenges the framework with dynamic lighting, diverse object shapes, and sizes. Our model demonstrates robustness in these conditions, accurately tracking and maintaining consistent IDs across different object types, from smaller items like a wok (ID 3) to larger potted plants (IDs 5 and 6). Based on the constructed 3D field, Ego3DT can stably track objects in outdoor environments using spatial relationships.

\paragraph{Indoor Persistence in Ego3DT-indoor.} Transitioning to the indoor domain, the Ego3DT-indoor dataset offers a contrasting setting with more controlled lighting but equally complex object interactions. The model successfully delineates and tracks objects such as furniture (IDs 1 to 4) in a typical room scenario, highlighting its precision in cluttered, confined spaces. The tracking continuity is evident, with the framework skillfully handling occlusions and varying distances from the camera.

\subsubsection{Qualitive Results on 2D Tracking.}
As shown in Figure~\ref{fig:case_track}, we compare our method to ByteTrack~\cite{zhang2022bytetrack} and our Ego3DT framework excels at preserving object identity across frames. The Ground Truth (a) provides the correct tracklets. Significant scene changes, like a person moving, require 3D scene information to track objects over time. ByteTrack coupled with GLEE detection (b) provides a strong baseline but occasionally falters with ID switches and detection lapses, especially under the erratic motion intrinsic to ego-centric videos. In contrast, our approach (c) demonstrates a remarkable grasp of object trajectories, maintaining accurate IDs even in the presence of motion blur and rapid scene changes. 

% As shown in Figure~\ref{fig:case_track}, we compare our method to ByteTrack~\cite{zhang2022bytetrack} and demonstrate how our approach offers improved detection stability. Our Ego3DT reliably identifies and tracks the objects. As seen in the side-by-side comparison of 2D tracking methods, our Ego3DT framework excels at preserving object identity across frames. The Ground Truth (a) provides a benchmark with flawless tracking and ID fidelity. Significant scene changes, like a person moving, require 3D scene information to track objects over time. ByteTrack coupled with GLEE detection (b) provides a strong baseline but occasionally falters with ID switches and detection lapses, especially under the erratic motion intrinsic to ego-centric videos. In contrast, our approach (c) consistently demonstrates a remarkable grasp of object trajectories, maintaining accurate IDs even in the presence of motion blur and rapid scene changes. These qualitative results show the advanced tracking performance of our Ego3DT.
% For example, Ego3DT can keep the consistency of the ID 45 with the extreme camera change. ByteTrack with GLEE detection has more inaccurate IDs and missing objects. These qualitative results show the advanced tracking performance of the proposed Ego3DT in dynamic ego-centric videos, highlighting its potential for comprehensive scene understanding and robust object tracking in diverse scenes. 

\section{Limitation, Future Work and Conclusion}
Although our proposed Ego3DT can successfully detect and track almost every 3D object in the scene, it might still fail in tracking some rapidly moving objects like cats, dogs, or humans. We leave this in future work, including tracking the moving objects and detecting the interaction with the scene and other objects. 

We introduce the Ego3DT framework for understanding RGB ego-centric video by leveraging 3D structure and ego motion to localize objects. Ego3DT utilizes existing 3D evaluators to construct 3D scenes based solely on RGB videos and achieves object recognition in both 2D and 3D through the open-vocabulary object detector and segmentor. Additionally, we build complete 3D scenes temporally using the sliding window mechanism and dynamic matching, enabling stable 2D object tracking by leveraging the constructed 3D positions. Our experimental results demonstrate that the Ego3DT framework outperforms existing methods, facilitating practical applications in augmented reality and robotics. 

% \section{Conclusion}
% We introduce the Ego3DT framework for understanding RGB ego-centric video by leveraging 3D structure and ego motion to localize objects. Ego3DT utilizes existing 3D evaluators to construct 3D scenes based solely on RGB videos and achieves object recognition in both 2D and 3D through the open-vocabulary object detector and segmentor. Additionally, we build complete 3D scenes temporally using the sliding window mechanism and dynamic matching, enabling stable 2D object tracking by leveraging the constructed 3D positions. Our experimental results demonstrate that the Ego3DT framework outperforms existing methods, facilitating practical applications in augmented reality and robotics. 

\section{Acknowledgments}

This work is supported by National Science Foundation of China under Grant 62106219, Zhejiang Provincial Natural Science Foundation of China under Grant LD24F020016 and LZ24F030005, and National Science Foundation for Distinguished Young Scholars under Grant 62225605.

\bibliographystyle{ACM-Reference-Format}
\balance
\bibliography{ref}

\end{document}